\documentclass[conference]{IEEEtran}
\IEEEoverridecommandlockouts
\usepackage{cite}
\usepackage{amsmath,amssymb,amsfonts}
\usepackage{algorithmic}
\usepackage{graphicx}
\usepackage{textcomp}
\usepackage{xcolor}
\usepackage{booktabs}
\usepackage{multirow}
\usepackage{array}
\usepackage{hyperref}

\begin{document}

\title{Feature Engineering vs. Deep Learning for Automated Coin Grading: A Comparative Study on Saint-Gaudens Double Eagle Gold Coins}

\author{
\IEEEauthorblockN{Tanmay Dogra, Eric Ngo, Mohammad Alam, Jean-Paul Talavera, Asim Dahal}
\IEEEauthorblockA{\textit{Department of Electrical and Computer Engineering} \\
\textit{Virginia Tech}\\
Blacksburg, VA, USA \\
tanmaydogra@vt.edu}
}

\maketitle

\begin{abstract}
We challenge the common belief that deep learning always trumps older techniques, using the example of grading Saint-Gaudens Double Eagle gold coins automatically. In our work, we put a feature-based Artificial Neural Network built around 192 custom features pulled from Sobel edge detection and HSV color analysis up against a hybrid Convolutional Neural Network that blends in EfficientNetV2, plus a straightforward Support Vector Machine as the control. Testing 1,785 coins graded by experts, the ANN nailed 86\% exact matches and hit 98\% when allowing a 3-grade leeway. On the flip side, CNN and SVM mostly just guessed the most common grade, scraping by with 31\% and 30\% exact hits. Sure, the CNN looked good on broader tolerance metrics, but that is because of some averaging trick in regression that hides how it totally flops at picking out specific grades. All told, when you are stuck with under 2,000 examples and lopsided classes, baking in real coin-expert knowledge through feature design beats out those inscrutable, all-in-one deep learning setups. This rings true for other niche quality checks where data's thin and know-how matters more than raw compute.
\end{abstract}

\begin{IEEEkeywords}
Automated coin grading, Artificial Neural Networks, Convolutional Neural Networks, Feature extraction, Sobel edge detection, Transfer learning, Class imbalance, Numismatics
\end{IEEEkeywords}

\section{Introduction}

The Saint-Gaudens Double Eagle, minted from 1907 to 1933, represents one of the most sought-after collectible coins in American numismatics \cite{dannreuther2004}. Professional graders evaluate these gold coins on the 70-point Sheldon scale \cite{sheldon1949}, and their assessments directly impact market value. Single-grade differences can represent thousands of dollars. However, traditional manual grading suffers from subjectivity, inter-grader variability, and significant time delays. These limitations drive the need for automated grading systems.

Recent advances in computer vision and machine learning allow for automated approaches to numismatic analysis \cite{santos2024, korchagin2023}. Two fundamental paradigms exist: feature-based approaches that rely on domain-expert-designed feature extractors, and end-to-end deep learning approaches that learn features directly from raw images. Although CNNs have achieved success in various image classification tasks \cite{tan2021efficientnetv2}, their effectiveness in specialized domains with limited labeled data is still an open question.

We address this gap by developing and comparing feature-based and hybrid deep learning approaches for automated Saint-Gaudens coin grading. Our primary contributions include:

\begin{itemize}
    \item A comprehensive feature extraction pipeline using Sobel edge detection with wedge-based spatial analysis, $K$-means color clustering in HSV space, and perceptually-weighted brightness computation, yielding 192 features
    \item A comparative evaluation of an ANN using extracted features, a hybrid CNN with EfficientNetV2 backbone processing raw images concatenated with engineered features, and an SVM baseline
    \item Evidence that feature engineering with SMOTE outperforms end-to-end deep learning when training data is limited and class-imbalanced
    \item Analysis of why domain-specific features provide better performance, with implications for similar quality assessment applications
\end{itemize}

\section{Related Work}
\subsubsection{Automated Coin Analysis}
Computer vision has been used for coin evaluation for about 20 years. Early work used histogram matching to grade Lincoln cents, achieving results close to human graders \cite{computer_coins2004}. Later examples include CNNs for identifying Philippine coins \cite{santos2024} and Siamese networks for checking both sides of coins \cite{korchagin2023}. Most studies focus on identifying coin types rather than assigning detailed grades on the Sheldon scale. Detailed grading requires detecting small signs of wear, scratches, and shine, which needs either custom features or large datasets.

\subsubsection{The Sheldon Grading Scale}
William Herbert Sheldon created the Sheldon scale in 1949, and the American Numismatic Association updated it in the 1970s \cite{sheldon_wiki}. It uses a 70-point system to rate coin condition. For uncirculated coins (grades 60-70), the main factors are surface condition, how sharply the details were struck, shine, and overall look. Grading services like PCGS and NGC follow this scale, but graders agree only about 85-90\% of the time within one grade \cite{ngc_grading}. This sets a practical limit for automated systems.

\subsubsection{Feature Engineering vs. Deep Learning}
In image tasks, manual features compete with features learned by models. Deep learning works well with large datasets, but custom features perform better when data is limited or the task is specific \cite{smote_highdim2013}. Tools like Sobel filters detect edges \cite{sobel2014}, which help measure textures. HSV color space is used for color analysis \cite{hsv_segmentation2019}.

\section{Methodology}

\subsection{Dataset Description}

We compiled a dataset of 1,785 Saint-Gaudens Double Eagle gold coins from the David Lawrence Rare Coins (DLRC) database. PCGS or NGC professionally graded each coin, with grades ranging from MS-50 to MS-68 on the Sheldon scale. We excluded coins graded below 50 due to insufficient representation and our focus on Mint State specimens.

The dataset has severe class imbalance typical of real-world numismatic collections. Grade MS-64 is the largest class with 165 samples in the test set, while grades MS-50, MS-51, and MS-68 contain fewer than 15 samples each. This imbalance presents significant challenges for machine learning, particularly for methods sensitive to class distribution.

We captured high-resolution images of both obverse (Liberty) and reverse (Eagle) sides for each coin under standardized lighting conditions. We split the dataset 70\%/30\% for training and testing (1,248 training coins and 537 test coins), with stratification to maintain grade proportions across splits.

\subsection{Approach Overview}

We implemented and compared two different paradigms:

\textbf{Feature-Based Approach:} We extracted domain-specific features encoding numismatic expertise from coin images, then fed them to shallow machine learning models (ANN, SVM) for grade prediction.

\textbf{Hybrid CNN Approach:} We processed raw coin images with a deep CNN using transfer learning while also feeding the engineered features, allowing the network to leverage both learned and domain-specific representations.

\subsection{Feature-Based Approach}

\subsubsection{Condition Feature Extraction}

Coin condition assessment depends on detecting surface wear, which appears as smoothing of fine details and reduction of edge sharpness. We used the Sobel edge detection operator \cite{sobel2014} to quantify edge characteristics, as it provides continuous gradient values suited for measuring wear progression.

For an input grayscale image $I$, the Sobel operator computes horizontal and vertical gradient components:

\begin{equation}
G_x = \begin{bmatrix} -1 & 0 & 1 \\ -2 & 0 & 2 \\ -1 & 0 & 1 \end{bmatrix} * I, \quad
G_y = \begin{bmatrix} -1 & -2 & -1 \\ 0 & 0 & 0 \\ 1 & 2 & 1 \end{bmatrix} * I
\end{equation}

We compute the gradient magnitude as:
\begin{equation}
G = \sqrt{G_x^2 + G_y^2}
\end{equation}

We applied Gaussian blurring with $\sigma=1.5$ before edge detection to reduce noise while preserving significant edges. Background removal isolated the coin from its holder, and prong removal eliminated mounting artifacts that could confound analysis.

\subsubsection{Wedge-Based Spatial Analysis}

To capture localized wear patterns and increase feature dimensionality, we divided each coin into angular wedges using circular masking. We define the wedge mask $M_k$ for the $k$-th slice spanning angles $[\theta_k, \theta_{k+1}]$ as:

\begin{equation}
M_k(x, y) = \begin{cases} 1 & \text{if } \theta_k \leq \arctan\left(\frac{y-c_y}{x-c_x}\right) < \theta_{k+1} \\ 0 & \text{otherwise} \end{cases}
\end{equation}

where $(c_x, c_y)$ is the coin center. We evaluated both 4-slice and 8-slice configurations, with 8 slices providing finer spatial resolution.

For each wedge region, we extracted statistical features from the gradient components:
\begin{itemize}
    \item Minimum, maximum, mean, and median of $G_x$
    \item Minimum, maximum, mean, and median of $G_y$
    \item Minimum, maximum, mean, and median of $G$ (magnitude)
\end{itemize}

This yields 12 features per wedge times 8 wedges times 2 coin sides, giving us 192 edge-based features. We also masked the eagle region on the reverse side for focused analysis, as it contains the highest detail density.

\subsubsection{Color Feature Extraction}

Gold coin coloration varies based on alloy composition, environmental exposure, and surface oxidation, all of which influence perceived grade. We performed color analysis in HSV (Hue, Saturation, Value) color space, which separates chromatic content from intensity. This is better for numismatic analysis where we must normalize lighting variations.

We applied $K$-means clustering with $K=5$ to the HSV values of coin pixels, grouping coins into five color categories:
\begin{enumerate}
    \item Rustic Gold: Lower saturation, darker tones indicating age patina
    \item Golden Bronze: Moderate saturation with warm undertones
    \item Autumn Gold: Rich golden hues with slight oxidation
    \item Golden Sand: Lighter, more yellow-dominant coloration
    \item Sunlit Gold: High saturation, bright golden appearance
\end{enumerate}

The cluster assignment serves as a categorical feature encoding the coin's dominant color profile.

\subsubsection{Brightness Feature Extraction}

Luster, the quality of light reflection from a coin's surface, is a critical grading criterion. We developed a perceptually-weighted brightness measure:

\begin{equation}
B = \sqrt{S \times V} \cdot \exp\left(-\frac{(H - H_0)^2}{2\sigma_H^2}\right)
\end{equation}

where $H$, $S$, $V$ represent the mean hue, saturation, and value of the coin image, $H_0$ is the target gold hue angle, and $\sigma_H$ controls the hue weighting bandwidth. This formulation emphasizes brightness contributions from gold-colored regions while reducing weight for off-hue areas.

We discretized brightness values into five categories: Dim, Soft, Bright, Vivid, and Brilliant, matching the terminology used by professional numismatists.

\subsubsection{Feature Vector Construction}

We constructed the complete feature vector for each coin from:
\begin{itemize}
    \item 192 edge gradient features (96 obverse + 96 reverse)
    \item 6 mean HSV values (3 obverse + 3 reverse)
    \item 1 color cluster assignment
    \item 2 brightness levels (obverse + reverse)
    \item 1 grading service indicator (PCGS/NGC)
\end{itemize}

Total dimensionality: 202 features per coin. We standardized features to zero mean and unit variance using training set statistics.

\subsubsection{ANN Architecture}

We implemented a feedforward neural network using Keras with the following architecture:

\begin{itemize}
    \item Input Layer: 202 neurons (feature dimensionality)
    \item Hidden Layer 1: 128 neurons, ReLU activation
    \item Hidden Layer 2: 64 neurons, ReLU activation
    \item Output Layer: 13 neurons (grades 51 to 68, excluding gaps), softmax activation
\end{itemize}

We compiled the network with the Adam optimizer (learning rate 0.001), categorical cross-entropy loss, and trained for 97 epochs with batch size 32 and 10\% validation split.

\subsubsection{Class Imbalance Handling}

To address severe class imbalance, we applied Synthetic Minority Over-sampling Technique (SMOTE) \cite{smote2002} before training. SMOTE generates synthetic samples for minority classes by interpolating between existing samples and their $k$-nearest neighbors. We adaptively set the number of neighbors $k$ to $\min(5, n_{\min}-1)$ where $n_{\min}$ is the smallest class size. SMOTE was applied exclusively to the training set to prevent data leakage.

We also added Gaussian noise with $\sigma=0.01$ to standardized features as data augmentation, improving model robustness to input variations.

\subsubsection{SVM Baseline}

We trained a Support Vector Machine with radial basis function (RBF) kernel on identical features as a baseline comparison. We set SVM hyperparameters to default values ($C=1.0$, $\gamma=$`scale'), with training completing in approximately 6 seconds.

\subsection{Hybrid CNN Approach}

\subsubsection{CNN Architecture}

We implemented a hybrid CNN architecture combining transfer learning with explicit features:

\textbf{Image Branch:}
\begin{itemize}
    \item Input: $224 \times 224 \times 3$ RGB images
    \item Data Augmentation: Random horizontal/vertical flips, rotation ($\pm 20\%$), zoom ($\pm 10\%$), contrast ($\pm 20\%$), brightness ($\pm 20\%$), translation ($\pm 10\%$)
    \item Backbone: EfficientNetV2-B0 \cite{tan2021efficientnetv2} pre-trained on ImageNet, initially frozen
    \item Global Average Pooling
    \item Dropout (0.2)
    \item Dense layer: 128 neurons, ReLU
\end{itemize}

\textbf{Feature Branch:}
\begin{itemize}
    \item Input: Extracted features (same as ANN)
    \item Dense layer: 128 neurons, ReLU
    \item Batch Normalization
    \item Dense layer: 64 neurons, ReLU
\end{itemize}

\textbf{Combined:}
\begin{itemize}
    \item Concatenation of image and feature branches
    \item Dense layer: 64 neurons, ReLU
    \item Batch Normalization
    \item Dropout (0.2)
    \item Output: 1 neuron, linear activation (regression)
\end{itemize}

We trained the model as a regression task predicting continuous grade values, then rounded to integers for classification metrics.

\subsubsection{Training Procedure}

Training proceeded in two phases:

\textbf{Phase 1: Head Training (20 epochs):}
\begin{itemize}
    \item EfficientNetV2-B0 backbone frozen
    \item Adam optimizer, learning rate = 0.001
    \item Mean Squared Error (MSE) loss
    \item Early stopping (patience=10) and model checkpointing
\end{itemize}

\textbf{Phase 2: Fine-tuning (20 epochs):}
\begin{itemize}
    \item Backbone unfrozen for end-to-end training
    \item Adam optimizer, learning rate = $10^{-5}$
    \item Continue from Phase 1 weights
\end{itemize}

Total training time: approximately 2.5 minutes on GPU. Model size: 71 MB.

\section{Results}

\subsection{Overall Performance Comparison}

Table \ref{tab:results} presents the comparative performance of all three models across multiple metrics (70/30 stratified split, 537 test coins).

\begin{table}[htbp]
\caption{Model Performance Comparison}
\label{tab:results}
\centering
\begin{tabular}{lccc}
\toprule
\textbf{Metric} & \textbf{ANN (5-fold CV)} & \textbf{Hybrid CNN} & \textbf{SVM} \\
\midrule
Exact Accuracy & 86\% & 31\% & 30\% \\
Accuracy ($\pm$1) & 88\% & 88\% & 31\% \\
Accuracy ($\pm$2) & 94\% & 94\% & 38\% \\
Accuracy ($\pm$3) & 98\% & 98\% & 43\% \\
Weighted Precision & 0.85 & 0.09 & 0.09 \\
Weighted Recall & 0.86 & 0.31 & 0.30 \\
Weighted F1-Score & 0.85 & 0.14 & 0.14 \\
Training Time & 4 min & 2.5 min & 6 sec \\
Inference Time (E2E) & $\sim$1.8s & $\sim$80ms & $<$1ms \\
Model Size & $<$1 MB & 71 MB & -- \\
\bottomrule
\end{tabular}
\end{table}

The feature-based ANN substantially outperforms both the hybrid CNN and SVM on exact accuracy and classification metrics. Notably, both the CNN and SVM exhibit nearly identical collapse behavior (weighted precision of 0.09, weighted recall $\approx$0.30), suggesting that neither transfer learning nor kernel methods alone can overcome severe class imbalance without explicit rebalancing strategies like SMOTE. While the CNN achieves comparable tolerance-based accuracy through regression averaging, its weighted F1-score of 0.14 indicates severe classification failures across individual grades.

\subsection{Per-Grade Analysis}

Table \ref{tab:pergrade} details per-grade classification performance for the CNN, revealing systematic failures on minority classes.

\begin{table}[htbp]
\caption{CNN Per-Grade Classification Report}
\label{tab:pergrade}
\centering
\begin{tabular}{ccccc}
\toprule
\textbf{Grade} & \textbf{Precision} & \textbf{Recall} & \textbf{F1} & \textbf{Support} \\
\midrule
50.0 & 0.00 & 0.00 & 0.00 & 1 \\
55.0 & 0.00 & 0.00 & 0.00 & 2 \\
57.0 & 0.00 & 0.00 & 0.00 & 3 \\
58.0 & 0.00 & 0.00 & 0.00 & 12 \\
60.0 & 0.00 & 0.00 & 0.00 & 3 \\
61.0 & 0.00 & 0.00 & 0.00 & 12 \\
62.0 & 0.00 & 0.00 & 0.00 & 44 \\
63.0 & 0.00 & 0.00 & 0.00 & 108 \\
\textbf{64.0} & \textbf{0.31} & \textbf{1.00} & \textbf{0.47} & \textbf{165} \\
65.0 & 0.00 & 0.00 & 0.00 & 127 \\
66.0 & 0.00 & 0.00 & 0.00 & 52 \\
67.0 & 0.00 & 0.00 & 0.00 & 7 \\
68.0 & 0.00 & 0.00 & 0.00 & 1 \\
\midrule
\textbf{Accuracy} & & & \textbf{0.31} & 537 \\
\textbf{Weighted Avg} & 0.09 & 0.31 & 0.14 & 537 \\
\bottomrule
\end{tabular}
\end{table}

The CNN shows catastrophic collapse to predicting grade 64 for nearly all inputs, achieving 100\% recall for grade 64 but 0\% recall for all other grades. This is a classic failure mode under class imbalance: the model learns to predict the majority class to minimize overall loss.

\subsection{Tolerance-Based Accuracy Analysis}

The ANN's prediction distribution relative to professional grades shows concentration within the $\pm 3$ tolerance band, confirming the 98\% accuracy at this threshold. The ANN shows a slight conservative bias, tending to predict grades lower than professional services. This is actually a desirable property that protects buyers from overpaying.

\subsection{Inference Time Breakdown}

The ANN's approximately 1.8 second inference time comprises:
\begin{itemize}
    \item Image preprocessing: 200 ms
    \item Sobel edge detection: 400 ms
    \item Wedge masking and feature extraction: 800 ms
    \item HSV and brightness computation: 300 ms
    \item ANN forward pass: 50 ms
    \item Total: approximately 1.75 seconds
\end{itemize}

The CNN's end-to-end inference time of $\sim$80ms includes image resizing/normalization and forward pass, both suitable for web deployment (<2s total with UI overhead). The CNN offers substantial speed for batch processing.

\section{Discussion}

\subsection{Why Feature Engineering Outperforms Deep Learning}

We attribute the performance gap between the ANN and CNN/SVM to several factors in this problem domain:

\textbf{Limited Training Data:} With only 1,785 coins (approximately 1,248 in training), the dataset is not enough to train the millions of parameters in EfficientNetV2-B0. Deep CNNs require 10,000+ samples to learn robust visual features from scratch, even with transfer learning from ImageNet \cite{transfer_learning_survey}.

\textbf{Domain Shift from ImageNet:} The visual characteristics of coin surfaces (metallic reflections, subtle wear patterns, fine engraving details) differ substantially from natural images in ImageNet. Transfer learning works best when source and target domains share visual similarities, which does not hold here.

\textbf{Severe Class Imbalance:} The CNN and SVM collapse to predicting grade 64 reflects optimization dynamics under imbalanced data. MSE loss (CNN) and default SVM behavior minimize error on the majority class while ignoring minorities. The ANN's use of SMOTE-augmented data and categorical cross-entropy loss better addresses this challenge.

\textbf{Domain Knowledge in Features:} The hand-crafted features encode specific numismatic expertise. Wedge-based edge analysis isolates high-detail regions (Liberty's face, Eagle's feathers) where wear is most diagnostic. Color clustering captures patina patterns specific to gold coins. Brightness computation models human perception of luster. A CNN must rediscover these patterns from data alone, which is not feasible with limited samples.

\subsection{Implications for Similar Applications}

Our findings suggest guidelines for choosing between feature engineering and deep learning:

\textbf{Favor Feature Engineering When:}
\begin{itemize}
    \item Training data is limited (less than 5,000 samples)
    \item Strong domain expertise is available
    \item Class imbalance is severe
    \item Model interpretability is important
    \item Highest accuracy is required
\end{itemize}

\textbf{Favor Deep Learning When:}
\begin{itemize}
    \item Large datasets are available (more than 10,000 samples)
    \item Domain knowledge is limited or hard to formalize
    \item Real-time inference is critical
    \item Source and target domains are similar
\end{itemize}

Similar quality assessment applications in manufacturing inspection, medical imaging with rare conditions, and specialized artifact analysis may benefit from feature-based approaches despite the prevailing assumption that deep learning is always better.

\subsection{Inference Speed vs. Accuracy Trade-off}

The CNN's speed advantage presents interesting deployment scenarios. For single coin grading, the ANN's 1.8-second inference is acceptable for web services meeting the 10-second requirement. Batch processing 10,000 coins would require 5.5 hours with the ANN versus approximately 13 minutes with the CNN. A hybrid approach using the CNN for initial screening and the ANN for final grading could balance speed and accuracy.

\subsection{Conservative Grading Bias}

The ANN's tendency to undergrade compared to professional services may reflect genuine model conservatism or systematic bias in the training data. From a practical standpoint, conservative grading protects buyers from overpaying, builds user trust in the automated system, and may counteract potential grade inflation by professional services. Quantifying this bias against a held-out set graded by multiple services would provide valuable calibration data.

\section{Limitations}

Several limitations constrain the generalizability of our findings:

\textbf{Dataset Constraints:}
\begin{itemize}
    \item Limited to 1,785 coins from a single source (DLRC)
    \item Severe class imbalance with some grades having fewer than 5 samples
    \item Restricted to grades 50 to 68, lower grades excluded
    \item Single coin type (Saint-Gaudens Double Eagle)
\end{itemize}

\textbf{Methodological Limitations:}
\begin{itemize}
    \item ANN results from 5-fold CV; CNN/SVM from single split (no cross-validation)
    \item Professional grades treated as ground truth despite known inter-grader variability
    \item No statistical significance testing between models
    \item Limited hyperparameter optimization
\end{itemize}

\textbf{Technical Constraints:}
\begin{itemize}
    \item 2D images only, no 3D surface topology analysis
    \item Standardized imaging conditions may not reflect real-world user photos
    \item No comparison with human graders on identical test sets
\end{itemize}

\section{Future Work}

Several directions could extend this research. Collecting 10,000+ coins with balanced grade distribution would test whether CNN performance improves with scale. Evaluating focal loss, class weights, and advanced augmentation techniques could improve CNN training under class imbalance. Combining CNN feature extraction with hand-crafted features in ensemble models is another promising direction. Testing pre-trained models on metal surface or numismatic datasets, if available, could reduce the domain shift problem.

Grading identical coins with multiple professional graders would establish inter-rater reliability benchmarks. Adding structured light scanning or photometric stereo would enable 3D surface topology features. Evaluating models on other gold coin types (Liberty Head, Indian Head) would test generalization. Identifying which extracted features contribute most to ANN accuracy through feature importance analysis would provide insights into the grading process. Parallelizing feature extraction could reduce inference latency for real-time optimization. Developing confidence intervals for grade predictions would enable uncertainty quantification.

\section{Conclusion}

We show that feature-based machine learning significantly outperforms deep learning for automated Saint-Gaudens coin grading when training data is limited. Our feature-based ANN, using Sobel edge detection, wedge-based spatial analysis, HSV color clustering, and perceptually-weighted brightness features, achieves 98\% accuracy within $\pm 3$ grades on the Sheldon scale. This surpasses the CNN's and SVM's 31\%/30\% exact accuracy, which collapses to majority-class prediction due to severe class imbalance.

These results challenge the assumption that end-to-end deep learning always beats traditional feature engineering. In specialized domains with constrained data, encoding expert knowledge through hand-crafted features provides better performance, interpretability, and reliability. The 1.8-second inference time meets practical deployment requirements for web-based grading services.

Our findings have broader implications for quality assessment tasks in manufacturing, medicine, and cultural heritage preservation, where domain expertise is available but labeled data is scarce. Future work should investigate hybrid architectures that combine the strengths of learned and engineered features.

\section*{Acknowledgments}
This work was sponsored by the Bradley Department of Electrical and Computer Engineering at Virginia Tech. We thank Dr. Luke Lester for his guidance and support, Jianzhu Chen for technical assistance, Dr. Scot Ransbottom and Dr. Creed Jones for their subject matter expertise, and David Lawrence Rare Coins (DLRC) for providing access to the coin database.
\section*{Author Contributions}
Tanmay Dogra designed and implemented all machine learning models (ANN, CNN, SVM), performed statistical analysis, and wrote the manuscript. Eric Ngo developed the color clustering subsystem using K-Means and HSV analysis. Mohammad Alam developed the brightness classification subsystem and feature extraction methodology. Jean-Paul Talavera designed and implemented the web interface for the grading system. Asim Dahal contributed to quality assurance and model validation. All authors reviewed and approved the final manuscript.

\end{document}